\documentclass[runningheads]{llncs}
\usepackage{subcaption}
\usepackage{graphicx}
\usepackage{amsmath}
\usepackage{amssymb}
\usepackage[ruled,vlined]{algorithm2e}
\usepackage[toc,page]{appendix}
\usepackage{url}

\begin{document}
\title{A Predictive Coding account for Chaotic Itinerancy \thanks{This work was funded by the CY Cergy-Paris University Foundation (Facebook grant) and partially by Labex MME-DII, France (ANR11-LBX-0023-01).}}
%
%
\author{Louis Annnabi \and
Alexandre Pitti \and
Mathias Quoy}
\authorrunning{L. Annabi et al.}
%
\institute{ETIS UMR 8051, CY University, ENSEA, CNRS \\
\email{\{firstname\}.\{lastname\}@ensea.fr}}
\maketitle              
\begin{abstract}

As a phenomenon in dynamical systems allowing autonomous switching between stable behaviors, chaotic itinerancy has gained interest in neurorobotics research. In this study, we draw a connection between this phenomenon and the predictive coding theory by showing how a recurrent neural network implementing predictive coding can generate neural trajectories similar to chaotic itinerancy in the presence of input noise. We propose two scenarios generating random and past-independent attractor switching trajectories using our model.

\end{abstract}
\section{Introduction}

Chaotic Itinerancy (CI) describes the behavior of large non-linear dynamical systems consisting in chaotic transitions between quasi-attractors \cite{Tsuda1991,Kaneko2003}. It was first observed in a model of optical turbulence \cite{Ikeda1989}, using globally coupled map in a chaotic system \cite{Kaneko1990} and in high dimensional neural networks \cite{Tsuda1991}. From a neuroscientific point of view, this phenomenon is interesting as such systems exhibit complex behaviors that usually require a hierarchical structure in neural networks. Studying CI could help better understanding the mechanisms responsible for the emergence of structure in large populations of neurons.

In cognitive neuroscience, it is believed that attractors or quasi-attractors could represent perceptual concepts or memories, and that cognitive processes such as memory retrieval or thinking would require neural trajectories transitioning between such attractors. CI is also gaining interest in neurorobotics, as it allows to design agents with the ability to autonomously switch between different behavioral patterns without any external commands. Several studies have tried to model CI with learned attractor patterns. \cite{Yamashita2008,Namikawa2011} propose a method where this functional structure emerges from a multiple-timescale RNN. Behavioral patterns are encoded in a rapidly varying recurrent population while another population with a longer time constant controls transitions between these patterns. \cite{Inoue2020} models CI, using reservoir computing techniques\cite{Lukosevicius2009}, with the interplay between an input RNN and a chaotic RNN where desired patterns have been learned with innate trajectory training \cite{Laje2013}. 

In this work, we try to model the attractor switching behavior of CI with a RNN implementation taking inspiration from the Predictive Coding (PC) theory. We propose a model performing random and past-independent transitions between stable and plastic limit-cycle attractors.

According to PC \cite{Rao1999,Clark2013}, the brain is hierarchically generating top-down predictions about its sensory states, and updating its internal states based on a bottom-up error signal originating from the sensory level. This view can be implemented by having the generative model intertwined with error neurons that propagate the information in a bottom-up manner through the hierarchy. An online computation of the error at each level of the generative model makes it possible to dynamically infer the hidden states, using only local update rules. The proposed model implements PC using the free-energy formulation \cite{Friston2009c}, providing a variational Bayes frame for the inference mechanisms.

We show how an RNN implementation based on PC can be trained to generate a repertoire of limit cycle attractor trajectories, and how adding noise into the neural dynamics causes random transitions between the learned patterns.

\section{Methods}

In this section, we present the proposed RNN model and the corresponding derivations for the free-energy. We then describe the two hypothesized situations in which our model could exhibit attractor transitions dynamics, that we label mode A and mode B.

\subsection{RNN model}

\begin{figure*}[!ht]
    \centering
    \includegraphics[width=\textwidth]{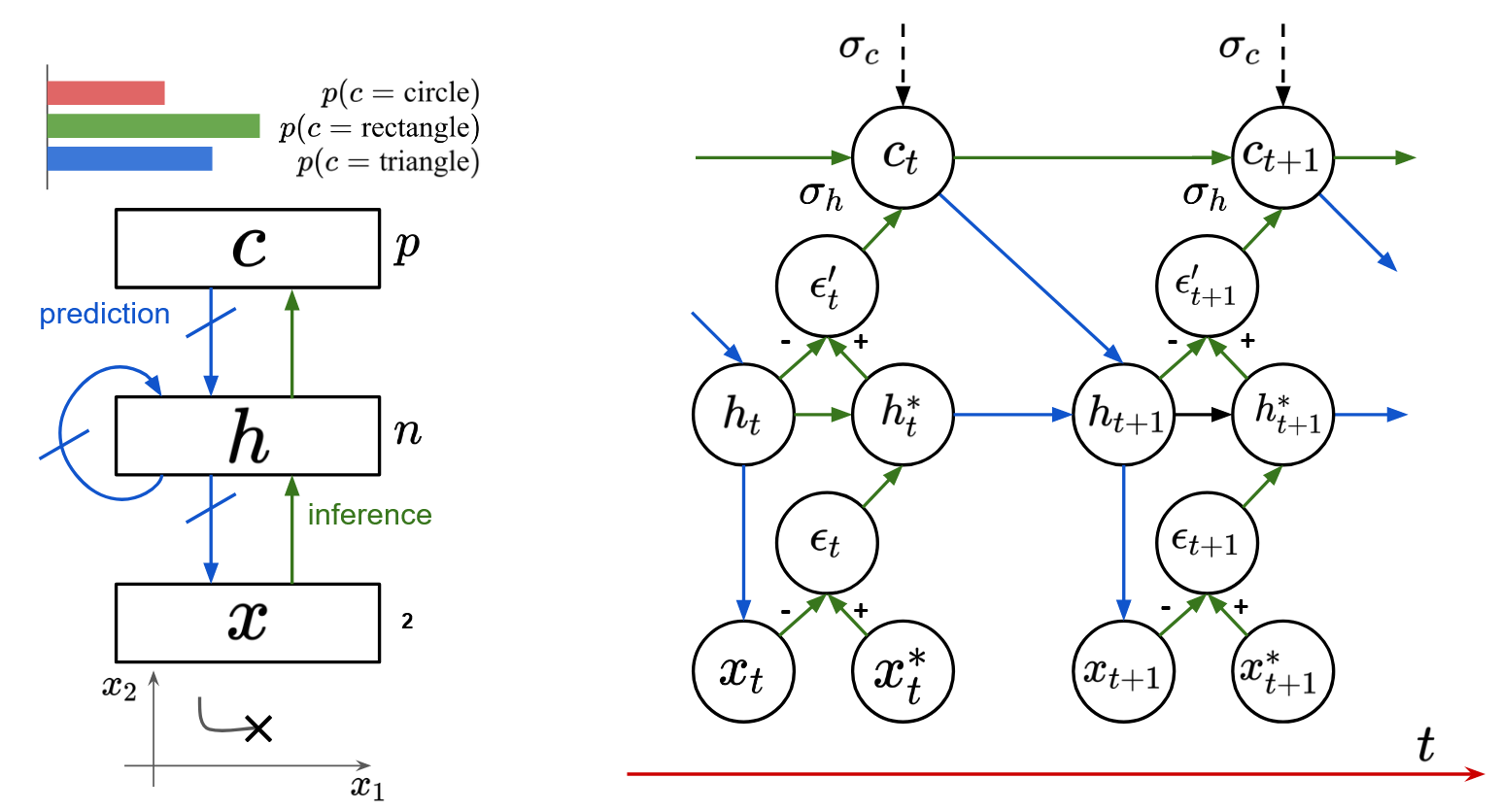}
    \caption{RNN model. Left: Functional block diagram of the model. The layers of the model interact through top-down connections (blue) and bottom-up connections (green). Right: Temporally unfolded computation graph of the model.}
    \label{fig:rnn_model}
\end{figure*}

Figure \ref{fig:rnn_model} represents our proposed RNN model implementing predictive coding. This implementation takes inspirations from several works on RNN modeling \cite{Ororbia2020,Taylor2009,Friston2009c}.

RNNs can be introduced as directed graphical models forming temporal sequences of hidden states $h_t$. RNNs can also include a sequence of input variables, and a sequence of output variables. The model we present here only considers outputs, that we denote $\mathbf{x}_t$. Such RNNs are parameterized by recurrent weights controlling the temporal evolution of $\mathbf{h}_t$, and output weights translating $\mathbf{h}_t$ into outputs $\mathbf{x}_t$.

Taking inspiration from \cite{Friston2009c}, we introduce hidden causes into our generative model. Hidden causes, that we denote $\mathbf{c}_t$, are variables influencing the temporal dynamics of $\mathbf{h}_t$. Contrary to hidden states, this variable is static and doesn't evolve according to recurrent weights. Hidden causes differ from model parameters, as they are a random variable on which we can perform inference. They also differ from inputs, as they are not an observable variable with known value. We still use the subscript $t$ on $\mathbf{c}_t$, since our model will perform inference at each time step, providing new estimates of the hidden causes variable.

To model the influence of the hidden causes variable $\mathbf{c}_t$ onto the temporal dynamics of the hidden states $\mathbf{h}_t$, we use a three-way tensor of shape $(n, n, p)$ where $n$ is the hidden state dimension and $p$ is the hidden causes dimension. The outcome of the dot product of this tensor by the hidden causes $\mathbf{c}_t$ is a matrix of shape $(n, n)$. We can thus see the three-way tensor as a basis of size $p$ in a dimensional space of recurrent weight matrices, and hidden causes as coordinates in this basis used to select particular temporal dynamics. Following this intuition that different hidden causes will lead to different hidden state dynamics, we choose to have one hidden causes vector for each attractor we want to learn with our model. To make sure these attractors don't interfere with each other during the training phase, we enforce one-hot embeddings for the hidden causes, with the activated neuron corresponding to the index of the attractor we want to learn. It ensues that the hidden causes dimension will be equal to the number of attractors we learn with this model.

This three-way tensor comprises a large number of parameters, causing this model to scale poorly if we increase the dimension of the hidden causes (i.e. the number of attractor patterns we learn). To address this issue, \cite{Taylor2009} proposes to factor the tensor into three matrices such that for all $i, j, k$, $\mathbf{W}_\mathbf{rec}^{ijk} = \sum_{l<d}\mathbf{W}_\mathbf{p}^{il} \cdot \mathbf{W}_\mathbf{f}^{jl} \cdot \mathbf{W}_\mathbf{c}^{kl}$. We introduce a factor dimension $d$ that we can be set arbitrarily to control the number of parameters. In our experiments, we used $d=n/2$.

The top-down, prediction pass through our network can thus be described with the following equations:

\begin{align}
    \mathbf{h}_t &= f(\mathbf{c}_{t-1}, \mathbf{h}_{t-1}^*) \\
    &=(1-\frac{1}{\tau}) \mathbf{h}^*_{t-1} + \frac{1}{\tau} \mathbf{W_f}
    \cdot ((\mathbf{W_c}^T \cdot \mathbf{c}_{t-1})(\mathbf{W_p}^T \cdot \tanh(\mathbf{h}_{t-1}^*))) \label{eq:h_pred}\\
    \mathbf{x}_t &= g(\mathbf{h}_t)\\
    &= \mathbf{W_{out}} \cdot \tanh(\mathbf{h}_t) \label{eq:x_pred}
\end{align}

Where we have introduced a time constant $\tau$ for the hidden state dynamics.

\subsection{Free-energy minimization}

As explained in introduction, our model implements PC with a bottom-up error propagation circuitry, represented with green lines in figure \ref{fig:rnn_model}. The error neurons, denoted $\mathbf{\epsilon}$ and $\mathbf{\epsilon'}$, compute the difference between predicted and target values at each layer. By propagating these errors originating from the output layer, onto the upper layers, this architecture is able to perform online inference of the hidden variables (states and causes) of the RNN.

Inference in the proposed model can be formulated as a free-energy minimization process. The detailed derivations of our model's equations based on the free-energy principle are provided in annex \ref{annex:fe_derivations}. We obtain the following equation for the free-energy $\mathcal{E}(\mathbf{h}, \mathbf{c})$:

\begin{equation}
    \mathcal{E}(\mathbf{h}, \mathbf{c}) 
    = \frac{(\mathbf{x}^* - \mathbf{x})^2}{2 \sigma_x^2}
    + \frac{(\mathbf{h}^* - \mathbf{h})^2}{2 \sigma_h^2}
    -  \log p(\mathbf{c})
    + C
\label{eq:fe}
\end{equation}

In this equation, $\mathbf{x}$ and $\mathbf{h}$ denote prior predictions while $\mathbf{h}^*$ denotes the approximate posterior estimation based on bottom-up information. $\mathbf{x^*}$ denotes the observed value. $C$ is a constant value that does not impact gradient calculations.

The probability $p(\mathbf{c})$ is the prior probability on the hidden causes variable. In this article, we use a Gaussian mixture prior, defined in the following equation:

\begin{equation}
    p(\mathbf{c}) = \sum_{k=1}^p \pi_k \mathcal{N}(\mathbf{c} ; \mathbf{\mu_k}, \sigma_c^2 \mathbb{I}_p)
\end{equation}

Note that the number of Gaussians in the mixture model is equal to $p$, which is the number of attractors, also equal to the dimension of $\mathbf{c}$.

The temporal dynamics of $\mathbf{h}$ and $\mathbf{c}$ can be found by computing the free-energy gradients with regard to these variables. The bottom-up, inference pass through our network is described by the following equations:

\begin{align}
    \mathbf{\epsilon}_t &= \mathbf{x}_t - \mathbf{x}_t^* \\
    \mathbf{h}_t^* &= \mathbf{h}_t - \frac{1}{\sigma_x^2}\mathbf{W_{out}}^T \cdot \mathbf{\epsilon_t} \label{eq:hidden_state_update}\\
    \mathbf{\epsilon}_t' &= \mathbf{h}_t - \mathbf{h}_t^* \\
    \mathbf{c}_t &= \mathbf{c}_{t-1} - \frac{1}{\sigma_h^2} \mathbf{W_c} \cdot ((\mathbf{W_f}^T \cdot \mathbf{\epsilon}_t')(\mathbf{W_p}^T \cdot \tanh(\mathbf{h}_{t-1}^*))) + \frac{\partial \log p(\mathbf{c_{t-1}})}{\partial \mathbf{c}_{t-1}} \label{eq:hidden_causes_update}
\end{align}

The last term in equation \ref{eq:hidden_causes_update} will pull $\mathbf{c}$ towards values with high prior probability.

Compared to the RNN proposed in \cite{Ororbia2020}, our model comprises hidden causes in the generative model. Additionally, the feedback connections perform gradient descent on the free-energy, instead of being additional parameters to be learned.

\subsection{Training}

\begin{algorithm}[!ht]
    \SetAlgoLined
    Initialize the RNN model\;
    $\mathbf{h}_{init} \sim \mathcal{N}(0, 1)$\;
    \For{$0 \leq i < I$}{
        \For{$0 \leq k < p$}{
            $\mathbf{h}_0 \gets \mathbf{h}_{init}$\;
            $\mathbf{c}_0 \gets$ one\_hot$(k)$\;
            $(\mathbf{x}_0, \dots, \mathbf{x}_T) \gets \text{RNN}(\mathbf{h}_0, \mathbf{c}_0)$\;
            $\mathcal{L} \gets \text{MSE}((\mathbf{x}_0, \dots, \mathbf{x}_T), (\mathbf{x}_0^*, \dots, \mathbf{x}_T^*))$\;
            backprop($\mathcal{L}$, RNN.parameters())\;
        }
    }
    \caption{RNN Training}
    \label{alg:training}
\end{algorithm}

The model can be trained with gradient descent on the free-energy functional using only local update rules. The output weights $\mathbf{W_{out}}$ can be trained in order to reduce the discrepancy between the observed value $\mathbf{x}_t^*$ and its prediction $\mathbf{x}_t$. Similarly, all the weights $\mathbf{W_p}$, $\mathbf{W_f}$ and $\mathbf{W_c}$, responsible for the temporal dynamics of $\mathbf{h}$, can be trained in order to reduce the error between the posterior estimation $\mathbf{h}_t^*$ and its prior estimation $\mathbf{h}_t$.

However, such learning rules would not consider the delayed influence of the recurrent weight parameters onto the trajectory. In this article, we instead use the backpropagation through time algorithm for the training of the model parameters, using only the forward pass described in equations (\ref{eq:h_pred}) and (\ref{eq:x_pred}) for gradient computations (all the bottom-up updates are detached from the computation graph).

For each limit cycle attractor $(\mathbf{x}^*_{0,k}, \mathbf{x}^*_{1,k}, \dots, \mathbf{x}^*_{T,k})$ of the $p$ trajectories we want to learn, we initialize the hidden causes to the one-hot encoding of $k$ (all coefficients set to 0 except for the $k$-th coefficient that is set to 1). All trajectories start from a same random initial hidden state $\mathbf{h}_{init}$. The training method is described in algorithm \ref{alg:training}.

Where $I$ denotes the number of training iterations, $T$ denotes the length of the target trajectories. During our training, we used the Adam optimizer with a learning rate of 0.01, and a batch size of $p$ corresponding to the inner loop in the previous algorithm.

\subsection{Mode A}

\begin{figure*}[!ht]
    \centering
    \begin{subfigure}{0.325\textwidth}
        \centering
        \includegraphics[width=\textwidth]{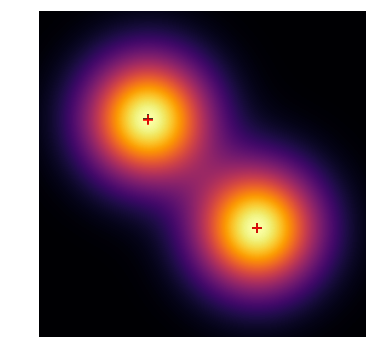}
        \caption{$\sigma_c = 0.4$}
        \label{fig:gmm_4}
    \end{subfigure}
    \begin{subfigure}{0.325\textwidth}
        \centering
        \includegraphics[width=\textwidth]{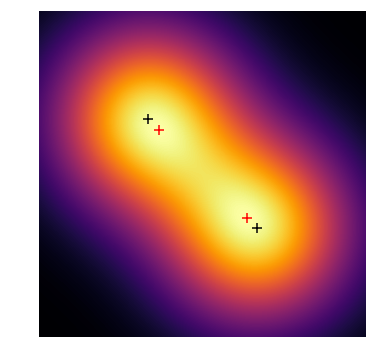}
        \caption{$\sigma_c = 0.6$}
        \label{fig:gmm_6}
    \end{subfigure}
    \begin{subfigure}{0.325\textwidth}
        \centering
        \includegraphics[width=\textwidth]{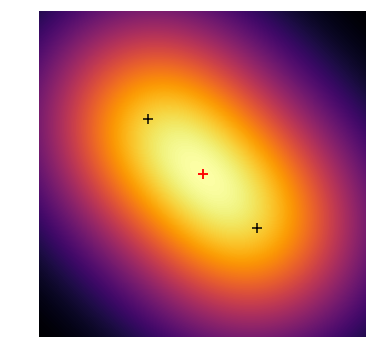}
        \caption{$\sigma_c = 0.8$}
        \label{fig:gmm_8}
    \end{subfigure}
    \caption{Gaussian mixture probability distributions with $p=2$. The Gaussians centers  $\mathbf{\mu}_0 = (1, 0)$ and $\mathbf{\mu}_1 = (0, 1)$ are represented in black. The red points represent the minima of the distributions. In the general case, the prior means $\mathbf{\mu}_k$ will correspond to the one-hot vectors activated on the $k$-th dimension, and the mixture coefficients $\pi_k$ will be set uniformly : $\pi_k = 1/p$.}
    \label{fig:gmm}
\end{figure*}

Here we describe one way to simulate attractor switching behavior using the proposed model. This method, that we label mode A, varies the parameters $\sigma_c$ used to dynamically infer hidden causes during the trajectory.

First, we are in a situation where no target $\mathbf{x}^*$ is provided by the environment, in other words, the RNN performs a closed-loop trajectory generation. In this situation, we replace the error in the bottom level by low amplitude noise. This noise propagates in the RNN with feedback connections and in particular, influences the hidden causes variable.

As represented in figure \ref{fig:gmm}, the parameter $\sigma_c$ determines the shape of the prior distribution on hidden causes. With low values of $\sigma_c$, the complexity term in equation (\ref{eq:hidden_causes_update}) will pull the hidden causes variable towards one of the prior means $\mathbf{\mu}_k$. These values for $\mathbf{c}$ correspond to temporal dynamics that have previously been trained to match each of the desired attractors. With high values of $\sigma_c$, the Gaussians merge into a concave function with a global maximum corresponding to the average of all the prior means $\mathbf{\mu}_{k}$. In this situation, the complexity term in equation (\ref{eq:hidden_causes_update}) will pull the hidden causes variable towards this average value, for which no training was performed.

The idea of mode A is to periodically vary $\sigma_c$ in order to alternate between phases where the hidden causes are pulled towards learned attractor dynamics values, and phases where the hidden causes are pulled towards the average of the prior means.

\subsection{Mode B}

We describe a second method to simulate attractor switching behaviors, that we label mode B. In mode B, the parameter $\sigma_c$ remains constant and equal to $0.4$, instead we vary the parameter $\sigma_h$. 

We can see from equation (\ref{eq:hidden_causes_update}) that this parameter controls the importance of the bottom-up signal in the hidden causes update. In our case, since the error that is propagated up into the model is pure noise, the parameter $\sigma_h$ can be seen as controlling the noise level that we add to the hidden causes at each time step. For high values of $\sigma_h$, the additive noise level will remain too low to pull the hidden causes outside of the basin of attraction created by the last term of equation (\ref{eq:hidden_causes_update}) and represented in figure \ref{fig:gmm_4}. For values of $\sigma_h$ that are low enough, the additive noise can make the hidden causes $\mathbf{c
}$ escape from its basin of attraction. 

Similarly to mode A, the idea behind mode B is to periodically vary $\sigma_h$ in order to alternate between low noise phases where hidden causes remain close to a value corresponding to the learned attractor dynamics, and high noise phases where the hidden causes escape their attraction basin.

\section{Results}

In this section, we present the results we obtained with the proposed model. We analyze the simulations of our network in mode A and mode B for the generation of attractor switching trajectories.

\subsection{Training}

We initialize our model with an output dimension of $2$, a hidden state dimension of $n=100$, and a hidden causes dimensions of $p=3$, equal to the number of attractor trajectories we want to learn. The network has a time constant of $\tau=5$. Finally, we set $\sigma_o=1$, $\sigma_h=10$ and $\sigma_c=0.1$ during training. Note that the parameters $\sigma_h$ and $\sigma_c$ will be varying during the simulations in mode A and B.

The three target trajectories are periodic patterns representing a circle, a square, and a triangle, with a period of 60 time steps, repeated to last for 1000 time steps.

The model was trained during 1000 iterations using the method described in Algorithm \ref{alg:training}.

\subsection{Mode A}

\begin{figure*}[!ht]
    \centering
    \includegraphics[width=1\textwidth]{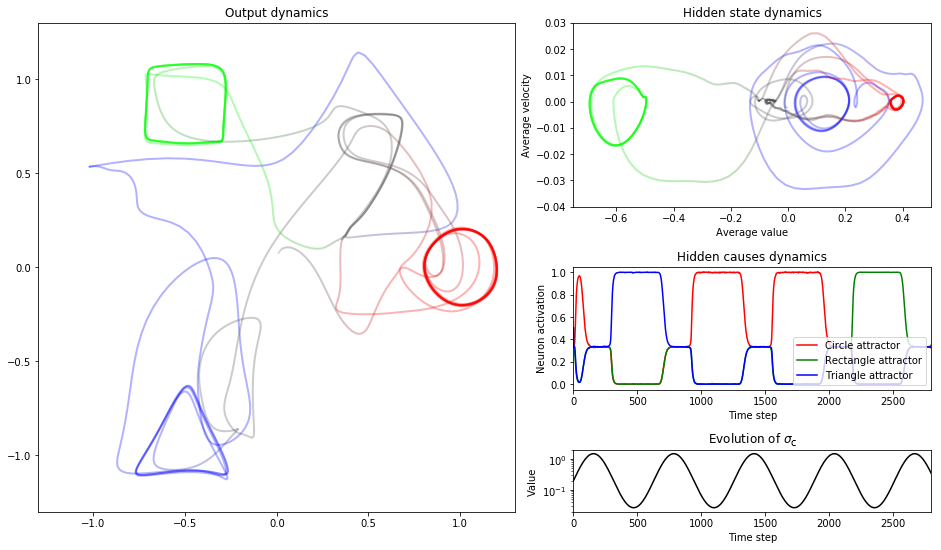}
    \caption{Simulation in mode A. Left: Output trajectory generated by the model in mode A. The line colors in RGB values correspond to the activations of the three neurons of $\mathbf{c}$ throughout the trajectory. Top-right: Average velocity of the hidden state according to its average value throughout the trajectory. Middle-right: Evolution of the three hidden causes neuron activations over time. Bottom-right: Evolution of the $\sigma_c$ coefficient over time.}
    \label{fig:ci_out_a}
\end{figure*}

We now use the trained network in mode A, with the parameters settings $\sigma_o = 10$, $\sigma_h=0.1$, and $\sigma_c$ varying according to the function $\sigma_c(t) = 0.2 * \exp\{2\sin(t/100)\}$. The results are recorded in figure \ref{fig:ci_out_a}.

We can observe that the RNN switches between the three attractors. When $\sigma_c$ is high, the hidden causes converge towards the center value. This center value corresponds to the hidden state dynamics and output dynamics depicted in gray. This value of the hidden causes seems to correspond to a point attractor, which was not something directly enforced by the training procedure. Starting from this configuration, when $\sigma_c$ decreases, the hidden causes falls into one of the three attracting configurations that were trained to correspond to the three limit cycle attractors.

\subsection{Mode B}

\begin{figure*}[!ht]
    \centering
    \includegraphics[width=1\textwidth]{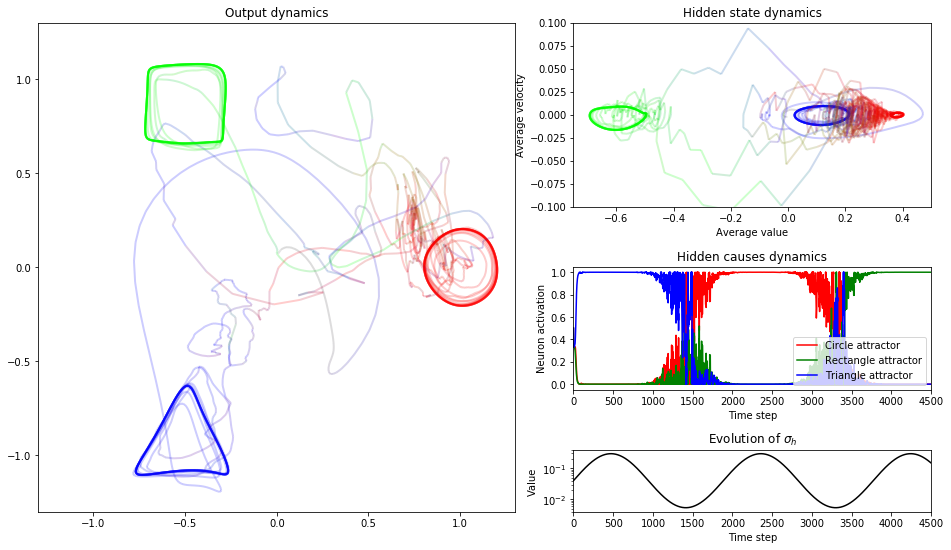}
    \caption{Simulation in mode B. Left: Output trajectory generated by the model in mode B. The line colors in RGB values correspond to the activations of the three neurons of $\mathbf{c}$ throughout the trajectory. Top-right: Average velocity of the hidden state according to its average value throughout the trajectory. Middle-right: Evolution of the three hidden causes neuron activations over time. Bottom-right: Evolution of the $\sigma_h$ coefficient over time.}    \label{fig:ci_out_b}
\end{figure*}

We now use the trained network in mode B, with the parameters settings $\sigma_o=10$, $\sigma_c=0.4$ and $\sigma_h$ varying according to the function $\sigma_h(t) = 0.04 * \exp\{2\sin(t/300)\}$. The results are recorded in figure \ref{fig:ci_out_b}.

We can observe that the RNN again switched between the three attractors. When $\sigma_h$ is high, the hidden causes remain in a stable position corresponding to the learned limit cycle attractor dynamics. When we decrease $\sigma_h$, the noise level applied onto the hidden causes at each time step increases to the point where $\mathbf{c}$ escapes its basin of attraction, to fall back into one of the three stable configuration once the noise level resettles.

\subsection{Transition matrices}

\begin{figure*}[!ht]
    \centering
    \begin{subfigure}{0.49\textwidth}
        \includegraphics[width=\textwidth]{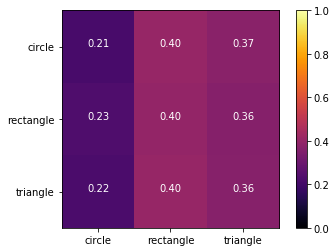}
        \caption{Transition matrix for mode A.}
        \label{fig:trans_a}
    \end{subfigure}
    \begin{subfigure}{0.49\textwidth}
        \includegraphics[width=\textwidth]{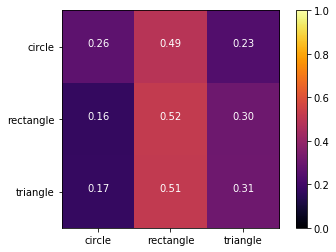}
        \caption{Transition matrix for mode B.}
        \label{fig:trans_b}
    \end{subfigure}
    \caption{Transitions matrices for modes A and B. Lines correspond to previous states and columns to next states. For instance, the estimated probability of switching from circle to triangle attractors in mode B is 0.23.}
    \label{fig:trans}
\end{figure*}

In this section, we want to verify whether the attractor switching behavior follows a uniform probability distribution or if some transitions are more likely to occur than others. We view the RNN as a Markov chain with three configurations. For modes A and B, we record 2000 attractor transitions that we use to build an estimation of the transition matrix of that Markov chain. The results are displayed in figure \ref{fig:trans}.

For mode A, we can see that the probability of switching to a certain state seems independent from the previous state. This result can be explained by the fact that the intermediary, neutral configuration that the networks reaches before switching to a new configuration corresponds to a fixed point. If we let enough time for the hidden state to reach this fixed point, it would no longer hold any memory of the previous configuration. Additionally, the probability distribution is not uniform, as rectangle states happen more often than others. 

For mode B, this bias is still present but contrary to mode A, the probability to reach a certain state depends on the previous state. The transitions are thus past-dependent.

\section{Conclusion}

In this study, we have shown how an RNN model implementing PC could exhibit attractor switching behaviors using an input noise signal. Here, we compare our results with other works aiming at modeling this behavior.

The approach described in \cite{Yamashita2008} requires to train a separate RNN for each primitive. In opposition, we have shown that our model can embed different dynamics within one RNN, and as such should scale better to an increased number of primitives. On the other hand, one limitation of the model presented by \cite{Inoue2020} is that quasi-attractors have a set duration, and the behaviour they yield can't last longer than this trained duration. In contrast, since our model relies on real trained limit-cycle attractors, any periodical behavior can be maintained for as long as desired.

In this article, we have tried to propose mechanisms that will provide random transitions between attractors, regardless of the past attractor state. However, if we were to model cognitive mechanisms such as memory retrieval, it could be interesting to have such a dependency. Following this idea, we could envision a mode C where we would periodically set the parameter $\sigma_c$ to a very large value. When $\sigma_c$ is very high, the prior probability over $\mathbf{c}$ converges to a flat function, thus making the last term of equation \ref{eq:hidden_causes_update} negligible. In such a setup, $\mathbf{c}$ would evolve following a Gaussian random walk. When $\sigma_c$ is reset to its initial value, $\mathbf{c}$ should converge to the closest mixture mean. Alternating between low values of $\sigma_c$ and very high values would thus result in a succession of random walk and convergence phases for $\mathbf{c}$, that should maintain information about the previously visited attractor configurations.

%

\bibliographystyle{splncs04}
\bibliography{bibliography.bib}

\newpage
\appendix
\section{Free-energy derivations}
\label{annex:fe_derivations}

In this section, we provide the derivations for equation \ref{eq:fe}. We start from the following probabilistic graphical model:

\begin{equation}
    p(\mathbf{c}) = \sum_{k=1}^p \pi_k \mathcal{N}(\mathbf{c} ; \mathbf{\mu_k}, \sigma_c^2 \mathbb{I}_p)
\end{equation}

\begin{equation}
    p(\mathbf{h}|\mathbf{c}) = \mathcal{N}(\mathbf{h} ; f(\mathbf{c}, \mathbf{h}_{t-1}); \sigma_h^2 \mathbb{I}_n)
\end{equation}

\begin{equation}
    p(\mathbf{x}|\mathbf{h}) = \mathcal{N}(\mathbf{x} ; g(\mathbf{h}); \sigma_x^2 \mathbb{I}_2)
\end{equation}

Where $f$ and $g$ correspond to the top-down predictions described respectively in equation \ref{eq:h_pred} and \ref{eq:x_pred}. Note that here, $\mathbf{c}$, $\mathbf{h}$ and $\mathbf{x}$ denote random variables, and should not be confused with the variables of the computation model presented in the main text. Since free-energy will be used to perform inference on the hidden variables, and that it's not possible to update the past hidden variable $\mathbf{h}_{t-1}$, we treat it as a parameter of function $f$ and only perform inference on $\mathbf{c}$ and $\mathbf{h} = \mathbf{h_t}$, where we have dropped the subscript. 

We introduce approximate posterior density functions $q(\mathbf{h})$ and $q(\mathbf{c})$ that are assumed to be Gaussian distributions of means $\mathbf{m}_h$ and $\mathbf{m}_c$. Given a target for $\mathbf{x}$, denoted $\mathbf{x^*}$, the variational free energy is defined as :

\begin{align}
    \mathcal{E}(\mathbf{x*}, \mathbf{m}_h,\mathbf{m}_c) &= \text{KL}(q(\mathbf{c}, \mathbf{h}) || p(\mathbf{c}, \mathbf{h}, \mathbf{x^*})) \\
    &= - \mathbb{E}_q[\log p(\mathbf{c}, \mathbf{h}, \mathbf{x^*})] + \mathbb{E}_q[\log q(\mathbf{c}, \mathbf{h})] \label{eq:fe_2}
\end{align}

The second term of equation \ref{eq:fe_2} is the entropy of the approximate posterior distribution, and using the Gaussian assumption, does not depend on $\mathbf{m}_h$ and $\mathbf{m}_c$. As such, this term is of no interest for the derivation of the update rule of $\mathbf{m}_h$ and $\mathbf{m}_c$, and is replaced by the constant $C_1$ in the remaining of the derivations. Using the Gaussian assumption, we can also find simplified derivations for the first term of equation \ref{eq:fe_2}, and grouping the terms not depending on $\mathbf{m}_h$ and $\mathbf{m}_c$ under the constant $C_2$, we have the following result:

\begin{align}
    \mathcal{E}(\mathbf{x*}, \mathbf{m}_h,\mathbf{m}_c) &= -\log p(\mathbf{x^*}|\mathbf{h}) - \log p(\mathbf{m}_h|\mathbf{c}) - \log p(\mathbf{m}_c) + C_1 + C_2 \\
    &= \frac{(\mathbf{x^*} - g(\mathbf{h}))^2}{2\sigma_x^2} + \frac{(\mathbf{m}_h - f(\mathbf{c}, \mathbf{h}_{t-1}))^2}{2\sigma_h^2}  - \log p(\mathbf{m}_c) + C
\end{align}

Where $C = C_1 + C_2 + C_3$ and $C_3$ corresponds to the additional terms obtained when developing $\log p(\mathbf{x^*}|\mathbf{h})$ and $\log p(\mathbf{m}_h|\mathbf{c})$.

\cite{Buckley2017} provides more detailed derivations and a deeper hindsight on the subject.

\section{Linked videos}

Here is the link to a video showing animated example trajectories in modes A and B (\url{https://youtu.be/LRJQr8RmeCY}).

\end{document}